\documentclass{article} 
\usepackage{iclr2026_conference,times}
\usepackage{tcolorbox}
\tcbuselibrary{minted,skins,breakable}
\usepackage{minted}
\usepackage{newtxtt}           

\newtcblisting{pybox}{
  listing engine=minted,
  minted language=python,
  minted options={
    breaklines,
    fontsize=\small,
    linenos,
    numbersep=6pt,
    autogobble,
    escapeinside=||
  },
  colback=gray!3,
  colframe=gray!40,
  arc=2pt, outer arc=2pt, boxrule=0.6pt,
  left=6pt,right=6pt,top=6pt,bottom=6pt,
  breakable,
  listing only   
}

\usepackage{tcolorbox}   
\tcbuselibrary{skins}    
\tcbuselibrary{breakable}
\usepackage{xcolor}      
\usepackage{tikz}        
\usepackage{varwidth}    
\usepackage{float} 

\usepackage[utf8]{inputenc}   
\usepackage[T1]{fontenc}      
\usepackage{times}             
\usepackage{inconsolata}       
\usepackage{microtype}         

\usepackage{amsmath, amssymb, amsfonts}
\usepackage{nicefrac}          

\usepackage{xcolor}            
\usepackage[table]{xcolor}     
\usepackage{colortbl}          
\usepackage{booktabs}          
\usepackage{multirow}          
\usepackage{makecell}          
\usepackage{array}             
\usepackage{adjustbox}         

\usepackage{graphicx}          
\usepackage{tikz}              
\usepackage{subcaption}        
\usepackage{float,wrapfig}     

\usepackage{hyperref}          
\usepackage{url}               

\usepackage{algorithm}
\usepackage{algorithmic}

\usepackage{pifont}            
\usepackage{xspace}            
\usepackage{amsthm}            
\usepackage{soul}              
\usepackage{ulem}              
\usepackage{caption}           
\usepackage{balance}           
\usepackage{natbib}

\usepackage{amsmath,amsfonts,bm}

\def\eqref#1{equation~\ref{#1}}

\def\1{\bm{1}}

\DeclareMathAlphabet{\mathsfit}{\encodingdefault}{\sfdefault}{m}{sl}
\SetMathAlphabet{\mathsfit}{bold}{\encodingdefault}{\sfdefault}{bx}{n}

\title{ContextNav: Towards Agentic Multimodal In-Context Learning}

\iclrfinalcopy
\author{Honghao Fu$^{1}$, Yuan Ouyang$^{2}$, Kai-Wei Chang$^{3}$, Yiwei Wang$^{4}$, Zi Huang$^{1}$, Yujun Cai$^{1}$\thanks{Corresponding Author} \\
$^1$The University of Queensland, $^2$Nanjing University, $^3$University of California, Los Angeles\\
$^4$University of California, Merced\\
\texttt{honghao.fu@uq.edu.au} \\
\url{https://contextnavpage.github.io/}\\
}

\begin{document}

\maketitle

\begin{abstract}
Recent advances demonstrate that multimodal large language models (MLLMs) exhibit strong multimodal in-context learning (ICL) capabilities, enabling them to adapt to novel vision-language tasks from a few contextual examples. However, existing ICL approaches face challenges in reconciling scalability with robustness across diverse tasks and noisy contextual examples: manually selecting examples produces clean contexts but is labor-intensive and task-specific, while similarity-based retrieval improves scalability but could introduce irrelevant or structurally inconsistent samples that degrade ICL performance. To address these limitations, we propose ContextNav, the first agentic framework that integrates the scalability of automated retrieval with the quality and adaptiveness of human-like curation, enabling noise-robust and dynamically optimized contextualization for multimodal ICL. {ContextNav unifies context management and noise-robust contextualization within a closed-loop workflow driven by graph-based orchestration.} Specifically, it builds a resource-aware multimodal embedding pipeline, maintains a retrievable vector database, and applies agentic retrieval and structural alignment to construct noise-resilient contexts. An Operational Grammar Graph (OGG) further supports adaptive {workflow planning} and optimization, enabling the agent to refine its {operational} strategies based on {downstream ICL} feedback. Experimental results demonstrate that ContextNav achieves state-of-the-art performance across various datasets, underscoring the promise of agentic workflows for advancing scalable and robust contextualization in multimodal ICL.
\end{abstract}

\section{Introduction}
\label{intro}

In-context learning (ICL) has emerged as a fundamental capability of large language models, enabling adaptation to novel tasks through contextual demonstrations without parameter updates~\citep{WMICLW}. By conditioning on task instructions and examples presented within the input context, ICL allows models to perform zero- or few-shot generalization without relying on gradient-based fine-tuning~\citep{LMFSL}. This paradigm has been successfully extended to multimodal domains, where models leverage both textual and visual examples to generalize across vision-language tasks~\citep{CoTcaption, MMICL, MemorizingMLM, TowardsMICL}.

\begin{figure*}[t]
    \centering
    \includegraphics[width=\linewidth]{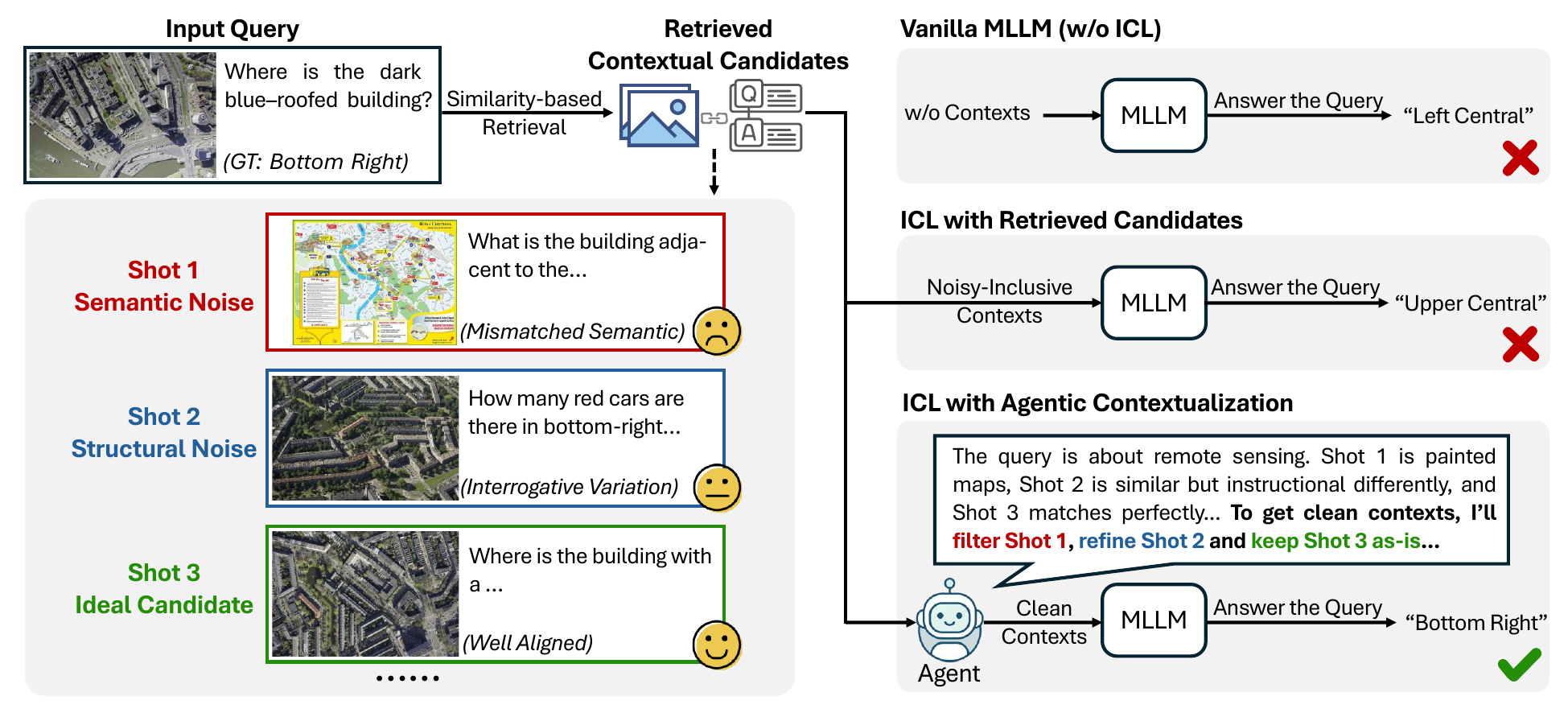}
    \caption{\small \textbf{Motivation for introducing agentic contextualization in multimodal ICL.} Similarity-based retrieval can introduce semantic or structural noise into contextual candidates, which degrades ICL effectiveness. Employing an agent for human-like curation could effectively alleviate this challenge. We provide a further discussion of the negative effects of such noise in Section~\ref{sec:dis}, grounded in quantitative results.}
    \label{fig:moti}
    \vspace{-0.14in}
\end{figure*}

Existing multimodal ICL methods can be broadly divided into two categories: Manual ICL, where examples are manually selected and organized into contexts~\citep{MultimodalCoT,TowardsUICL,TowardsMICL}, and Retrieval-based ICL, which employs feature embeddings to retrieve candidate examples as contexts~\citep{MakesICL,VPSICLS,PMICL}. While both approaches have shown promising results, they face notable challenges. Manual ICL often yields highly relevant and well-structured contexts but relies heavily on human curation, making it labor-intensive and difficult to generalize across large-scale multimodal corpora. {Retrieval-based ICL alleviates this burden through automation; however, it may also retrieve semantically irrelevant samples and samples with inconsistent interrogative, imperative, or narrative structures, both of which can degrade downstream ICL performance.} As illustrated in Figure~\ref{fig:moti}, such noisy contexts can mislead models into producing incorrect answers.

Moreover, retrieval remains a static and rule-based one-shot process. Unlike human reasoning that {adaptively} refines example selection based on observed effectiveness, current systems cannot adapt their contextualization strategies or learn from experience. This creates a fundamental challenge: developing contextualization approaches that combine the scalability of automated retrieval with the quality and adaptiveness of human-like curation.

To address these challenges, we introduce ContextNav, {an agentic framework for contextualization in multimodal ICL.} Unlike previous methods, ContextNav formulates contextualization as an adaptive, tool-driven, and automated workflow that systematically manages multimodal context corpora, filters out noisy examples, reorganizes retrieved candidates in a human-like manner, and optimizes its tool orchestration strategies based on downstream ICL feedback. Specifically, ContextNav leverages the multimodal reasoning capability of the MLLM policy to build and maintain a resource-aware embedding pipeline with a retrievable vector database, and applies agentic retrieval and structural alignment to construct noise-resilient contexts. An Operational Grammar Graph (OGG) further supports adaptive {workflow} planning and optimization, enabling the agent to refine its {operational} strategies across multiple timesteps based on downstream ICL feedback. Collectively, these components constitute a {self-optimized} closed-loop system that not only retrieves and manages contexts, but also adaptively organizes them, thereby enabling automated and robust multimodal ICL. ContextNav eliminates manual contextualization, achieving greater scalability than manual methods, while its human-like curation confers stronger noise robustness than retrieval-based approaches. To the best of our knowledge, ContextNav is the first agentic framework for contextualization in multimodal ICL. Our contributions are summarized as follows:
\begin{itemize}
    \item We propose ContextNav, the first agentic framework that formulates multimodal contextualization as an adaptive, tool-driven, and fully automated agentic workflow, supporting scalable and noise-robust contextualization for multimodal ICL.
    
\item {ContextNav transcends static one-shot retrieval by integrating the Operational Grammar Graph with a memory module that couples historical {workflows} and downstream ICL feedback, enabling {adaptive optimization of workflow orchestration} across timesteps and enhancing adaptability and robustness of contextualization.}
    
    \item Through extensive experiments on diverse datasets, ContextNav achieves an average ICL gain of 16.8\% across models, surpassing the prior state-of-the-art (7.6\%) and underscoring the promise of agentic workflows for multimodal ICL.
\end{itemize}


\section{Related Works}
\noindent\textbf{In-Context Learning (ICL).} The concept of ICL was popularized by GPT-3~\citep{LMFSL} and has since naturally emerged as a prominent paradigm in natural language processing (NLP)~\citep{SurveyICL}. A growing body of work has sought to understand the underlying mechanisms of ICL in LLM, primarily attributing this capability to implicit gradient descent~\citep{G1,G2,G3} and Bayesian modeling frameworks~\citep{BayesianICL22,BayesianICL24,B1,B2}. 
Concurrently, numerous studies have also focused on enhancing the ICL capabilities of LLMs by improving their inference frameworks~\citep{PaLM,FADSICL,IFTB}, training strategies~\citep{Maml-en-llm,WhyICL}, and contextualization methods~\citep{Auto-icl,ELICIT,ICV}, enabling more effective ICL on downstream tasks~\citep{LIFT}. These efforts have advanced both the theoretical understanding and practical application of ICL.

\noindent\textbf{Multimodal ICL.} The success of ICL in language models has spurred growing research interest in extending this paradigm to multimodal domains~\citep{FactorsICL}. Early MLLMs, such as Flamingo~\citep{Flamingo}, InstructBLIP~\citep{InsBLIP}, and LLaVA~\citep{LlaVA}, have demonstrated the potential of multimodal ICL. More recent studies further advance this capability by incorporating multimodal chain-of-thought reasoning~\citep{MultimodalCoT}, vision expert models~\citep{TowardsUICL}, feature-based retrieval~\citep{MakesICL,improvingICL,MMHQA-ICL,LCICL}, in-context tuning~\citep{ICtuning}, representation engineering~\citep{MMICL,m2iv,MTV}, and attention editing~\citep{CAMA}. These techniques have collectively enhanced the zero- and few-shot ICL performance of VLMs across both specialized~\citep{CoTcaption,ICLvec} and general vision-language tasks~\citep{CanMLLMs}. Despite these advances, current approaches still face challenges from contextual noise. In this paper, we address contextual noise by proposing an agentic workflow that combines the scalability of automated retrieval with the quality and adaptability of human-like curation, thereby enabling noise-robust and dynamically optimized contextualization.




\section{ContextNav}
\subsection{Overview}

\begin{figure*}[t]
    \centering
    \includegraphics[width=\linewidth]{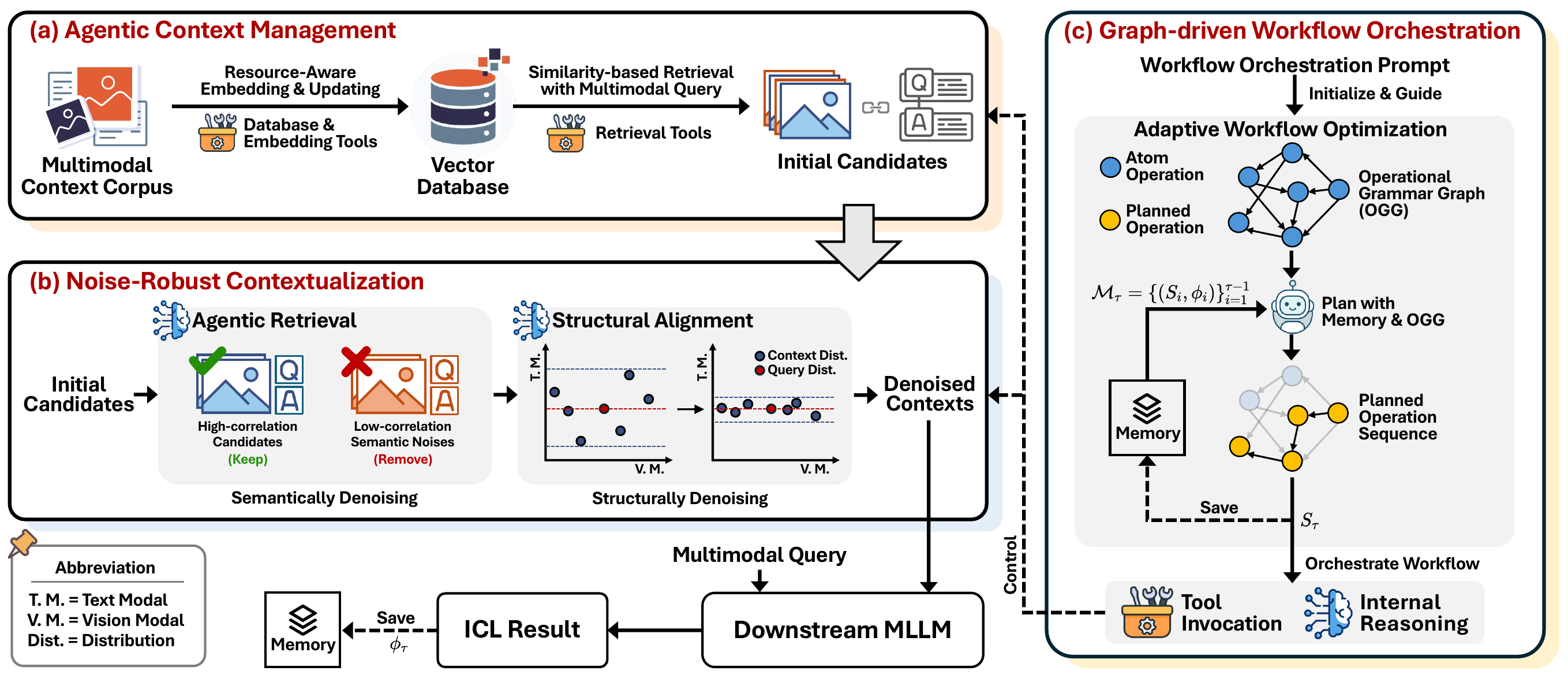}
    \caption{\small \textbf{Framework of the ContextNav.} The proposed agentic framework integrates three synergistic modules: (a) \textbf{Agentic Context Management}, which performs resource-aware multimodal embedding, maintains a continuously updated vector database, and subsequently leverages it for similarity-based retrieval to generate the initial candidate pool given an input query; (b) \textbf{Noise-Robust Contextualization}, which refines retrieved candidates through agentic retrieval and structural alignment to mitigate both semantic and structural noise; and (c) \textbf{Graph-driven Workflow Orchestration}, where the agent leverages an Operational Grammar Graph and memory module to {adaptively plan and optimize operation sequence, thereby controlling the workflow.} Collectively, these modules enable systematic management, representation, retrieval, and organization of multimodal contexts, supporting noise-robust and dynamically optimized contextualization for multimodal ICL.}
    \label{fig:pipe}
    \vspace{-0.1in}
\end{figure*}

We propose ContextNav, an agentic framework designed to advance the multimodal ICL performance of downstream MLLMs. Given a multimodal query as input, ContextNav establishes an end-to-end agentic pipeline that autonomously transforms raw corpora into well-formed, query-relevant contexts. As illustrated in Figure~\ref{fig:pipe}, the framework unfolds in three synergistic modules. First, the \textbf{Agentic Context Management} module constitutes the entry point. The agent performs resource-aware multimodal embedding, builds an evolving vector database, and retrieves {a group of initial candidates} from it given an input query. Second, the {resulting candidates are} feed into the \textbf{Noise-Robust Contextualization} module, where semantically and structurally noisy candidates are pruned or reorganized to yield cleaner contexts. In parallel, the \textbf{Graph-driven Workflow Orchestration} module oversees and coordinates these processes, ensuring that embedding, retrieval, and denoising operations form valid and {optimized operation sequences}. Collectively, these components establish an automated workflow for representing, managing, retrieving, and refining multimodal contexts, thereby supporting scalable and noise-resilient multimodal ICL for downstream MLLMs.

\subsection{Agentic Context Management}
\label{sec:acm}
\textbf{Resource-Aware Multimodal Embedding.} {Embedding multimodal corpora is a prerequisite for building vector databases and enabling effective retrieval, while it faces several challenges. Large-scale embedding incurs heavy computational and storage costs, often becoming a system bottleneck. Embedding models also vary in accuracy and efficiency, creating trade-offs between fidelity and resource use, as shown in the experimental results of Table~\ref{tab:embedding_models}. Moreover, {resource usage preference} differs across users, requiring adaptive allocation. These factors motivate a resource-aware design that balances performance with efficiency in a demand-driven manner. {To this end, ContextNav formulates multimodal embedding as an agent-driven process. The embedding-specification prompt $\mathcal{P}\text{emb}$ (detailed in Appendix~\ref{appendix_b} and Appendix~\ref{appendix_d}) encodes the user’s resource usage preferences, the current hardware state, and the embedding model zoo. Based on this prompt, the agent’s MLLM policy $\pi_{\theta}$ employs its internal reasoning capability to perform resource-aware hardware-model matching, determining an appropriate pair of embedding models $(E_T, E_V) = \pi_{\theta}(\mathcal{P}_\text{emb})$, where $E_T$, $E_V$ denote textual and visual embedding models, respectively.} Let the context corpus be denoted as {$\mathcal{C} = \{(T_i, I_i)\}_{i=1}^N$, where $T_i$ is the $i$-th text instance and $I_i$ is its paired image. The agent leverages database and embedding tools to instantiate the embedding process at timestep $\tau$, which can be expressed as:}
\begin{equation}
\mathcal{E}_\tau = \left\{ \big(E_\text{text}(T_i),\, E_\text{vis}(I_i)\big) \;\middle|\; (T_i, I_i) \in \mathcal{C}_\tau \right\},
\end{equation}
where $\mathcal{E}_\tau$ denotes the multimodal embedding set constructed from the context corpus $\mathcal{C}_\tau$.
In parallel, the agent employs {database tools} to continuously monitor the context corpus to identify newly added or modified samples $\Delta \mathcal{C}_{\tau+1}$ that have not been vectorized, thereby triggering an on-demand embedding pipeline for these samples in the subsequent timestep and yielding the corresponding embedding set $\Delta \mathcal{E}_{\tau+1}$. This ensures the database remains up-to-date.

\textbf{Vector Database Management.} Following the generation of embeddings, the agent undertakes the construction and adaptive management of a multimodal vector database {$\mathcal{D} = \left\{ \big(T_i, I_i, e_i\big) \;\middle|\; (T_i, I_i) \in \mathcal{C},\, e_i \in \mathcal{E} \right\}$.} Contexts and their representations are systematically ingested, indexed, and archived, thereby establishing a structured and readily retrievable database. To accommodate corpus dynamics, the agent maintains an adaptive update mechanism with {database tools}:
{
\begin{equation}
\mathcal{D}_{\tau+1} = \mathcal{D}_{\tau} \cup 
\left\{ \big(T_j, I_j, \mathbf{e}_j\big) \;\middle|\; (T_j, I_j) \in \Delta\mathcal{C}_{\tau+1},\, \mathbf{e}_j \in \Delta \mathcal{E}_{\tau+1} \right\},
\end{equation}}
This agentic orchestration transforms the database from a static collection into an evolving knowledge structure that accurately reflects the state of the underlying context corpus.

\textbf{Initial Candidate Pool.} Once the vector database is constructed, {the agent instantiates a Top-$k$ similarity-based retrieval function $f_\tau$, which is an adaptive composition of heterogeneous multimodal retrieval tools (textual, visual, or their cascaded combination).} Given a budget $k$, the function yields the initial candidate pool:
\begin{equation}
R_\tau^{init} \;=\; f_\tau\!\big(q,\, \mathcal{D}_\tau,\, k\big),
\end{equation}
where $q=(q_t,q_v)$ denotes a multimodal query composed of the textual query $q_t$ and its paired visual query $q_v$. {The generated initial candidate pool is subsequently used for contextualization.}

\subsection{Noise-robust Contextualization}
{Within ContextNav, agentic retrieval and structural alignment are employed to attenuate semantic and structural noise from the initial candidates, respectively, with both processes grounded in the internal reasoning of the MLLM policy. Semantic noise refers to candidates whose content is off-topic or contradictory to the query intent, whereas structural noise arises when a candidate’s interrogative, imperative or narrative structure diverges from that of the query.} 

\textbf{Agentic Retrieval.} As discussed in Section~\ref{intro}, {initial candidates} generated via similarity-based retrieval may contain both semantic and structural noise, which can undermine the downstream ICL performance. To mitigate the semantic noise introduced by weakly aligned matches, the agent subsequently applies a second-stage filtering beyond raw similarity search. Specifically, the coherence-specification prompt $P_{\text{coh}}$ (detailed in the Appendix~\ref{appendix_b}) encodes explicit instructions for semantic assessment, such as verifying topical consistency between query and candidate, and discarding candidates that contradict or distract from the query intent. Conditioned on this prompt, the MLLM policy $\pi_\theta$ leverages its internal reasoning capacity to evaluate each multimodal candidate in $R^{\text{init}}_{\tau}$ and decide whether to retain it. The process at timestep $\tau$ can be expressed as:
\begin{equation}
R_\tau^{sem} \;=\; \pi_\theta\!\big(q, \mathcal{P}_{\text{coh}},\, R_\tau^{init}\big).
\end{equation}
Through this agentic retrieval mechanism, the initial candidates are effectively denoised, ensuring the preservation of only those contexts that exhibit strong semantic alignment with the query.

\textbf{Structural Alignment.} 
{Contextual candidates may exhibit variability in interrogative, imperative or narrative structure.} Such heterogeneity introduces structural noise into the context, which prior work has extensively demonstrated to hinder consistent reasoning in ICL and lead to performance degradation~\citep{MindFormat,PromptConsistency,calibrate}. To mitigate this, the agent applies a structure-refinement step that harmonizes the form of retrieved candidates with the query. Specifically, the structure-alignment prompt $\mathcal{P}_{\text{str}}$ (detailed in the Appendix~\ref{appendix_b}) encodes explicit instructions for {reorganizing}, ensuring that the textual flow mirrors that of the input query $q_t$. Conditioned on this prompt, the MLLM policy $\pi_\theta$ leverages its {internal} reasoning capability to {edit} candidates with {divergent textual structures} into a form consistent with $q_t$, thereby reducing structural discrepancies. The process at timestep $\tau$ can be expressed as:
\begin{equation}
R_\tau^{alin} \;=\; \pi_\theta\!\big(q_t,\, \mathcal{P}_{\text{str}},\, R_\tau^{sem}\big).
\end{equation}
This process aligns the {textual structures of the candidates} with that of the query, reducing distributional bias and structural noise. By combining semantic denoising and structural alignment, the agent yields a noise-minimized context set that enhances the robustness of contextualization.}

\subsection{Graph-driven {Workflow} Orchestration}
\textbf{Operational Grammar Graph (OGG).} {In the proposed framework, the operations (tool invocation or internal reasoning) that the agent perform for context management and contextualization are governed by strict dependency relations and compositional constraints, since the underlying data structures evolve progressively within the workflow, with each stage reshaping the results of the previous one.}
Naive concatenation or heuristic composition risks redundant or invalid executions. Recent studies~\citep{langgraph, AGORA} demonstrate that graph structures effectively capture operational dependencies and control flows while enabling flexible multi-step workflows. Building on this insight, we construct a directed graph {$\mathcal{G}=(\mathcal{V},\mathcal{E})$}, termed OGG, where {$\mathcal{V}$} denotes the set of atomic {operations} and {$\mathcal{E}$} encodes valid execution dependencies, thereby formalizing the grammar of permissible operations (detailed in the Appendix~\ref{appendix_c}). {Specifically, the agent can instantiate a workflow that follows the operation sequence $S=(v_1 \to v_2 \to \cdots \to v_m)$, where each transition satisfies $(v_i, v_{i+1}) \in \mathcal{E}$, thus ensuring that the modules are executed in a valid order.}

\textbf{Adaptive {Workflow} Optimization.} {The agent plans and orchestrates workflows by exploiting the prior dependency structures encoded within the OGG $\mathcal{G}$.} However, under the constraint of one-shot planning without intermediate feedback, the instantiated workflow may be suboptimal, potentially introducing inaccurate contextual candidates. {This can reduce the number of valid contexts, shorten the effective context length, and ultimately degrade ICL performance, as illustrated in Figure~\ref{fig:shots}.} To address this, the agent adopts an {adaptive} optimization mechanism that leverages correlations between past workflow configurations and their downstream ICL performance stored in the memory $\mathcal{M}$. Conditioned on a workflow orchestration prompt $\mathcal{P}_{\text{wop}}$ (detailed in Appendix~\ref{appendix_b}), workflow instantiation at timestep $\tau$ is modeled as the operation sequence:
\begin{equation}
S_\tau \;=\; \pi_\theta\!\big(\mathcal{P}_{\text{wop}},\, \mathcal{M}_\tau,\, \mathcal{G}\big),
\end{equation}
{where $\mathcal{P}_{\text{wop}}$ specifies the workflow requirements at the initial timestep in the absence of downstream ICL feedback, and further declares the optimization logic governing subsequent timesteps. Notably, for operations that involve internal reasoning, the associated prompts (e.g., $\mathcal{P}_{\text{emb}}$, $\mathcal{P}_{\text{coh}}$, $\mathcal{P}_{\text{str}}$) are embedded internally within the operation itself rather than being directly specified in $\mathcal{P}{\text{wop}}$. This iterative optimization design enables the agent not only to leverage the OGG} to enforce execution validity but also to {adaptively} refine orchestration strategies across multiple timesteps, thereby integrating context management and noise-robust contextualization into a coherent and adaptive pipeline.

\subsection{Agentic Multimodal In-Context Learning}
At each timestep $\tau$, the agent constructs a noise-minimized context set $R_\tau^{alin}$, which is concatenated with the multimodal query $q$ and fed into the {downstream} MLLM $\Phi$. The execution of in-context learning is guided by the prompt $\mathcal{P}_{\text{icl}}$ (detailed in the Appendix~\ref{appendix_b}). The model then produces both the final prediction $y_\tau$ for the input query and an auxiliary textual feedback  $\phi_\tau$ that reflects the perceived quality of the constructed context. Formally, this process can be expressed as:
\begin{equation}
(y_\tau,\, \phi_\tau) \;=\; \Phi\!\big( R_\tau^{alin},\, q,\, \mathcal{P}_{\text{icl}} \big),
\end{equation}
The feedback $\phi_\tau$ provides an immediate assessment of context adequacy from the perspective of the MLLM performing ICL, thereby assisting the agent in optimizing its toolchain selection strategy. Specifically, the agent updates its memory $\mathcal{M}$ by recording the association between the executed operation sequence $S_\tau$ and the resulting feedback $\phi_\tau$:
\begin{equation}
\mathcal{M}_{\tau+1} \;=\; \mathcal{M}_{\tau} \cup \big\{ (S_{\tau}, \phi_{\tau}) \big\}= \{ (S_i, \phi_i) \}_{i=1}^{\tau}.
\end{equation}
This continual feedback update closes the loop between multimodal ICL and adaptive toolchain optimization, enabling the agent to progressively refine its planning strategy to select toolchains that provide stronger contextual support and thereby enhance the robustness of multimodal ICL.

\section{Experiment}

\subsection{Dataset and Implementation}
\label{sec:dataset}
\noindent\textbf{Dataset.} We first conduct a difficulty annotation of query samples from recent composite-task datasets and benchmarks, including BlindTest~\citep{blindtest}, MME-RealWorld~\citep{RealWorldBench}, CharXiv~\citep{CharXiv}, GVL~\citep{GVL}, and MathVision~\citep{mathvision}. The annotation is determined by the accuracy of the models under evaluation: if more than half of the tested models answer a given query correctly, it is labeled as easy; otherwise, it is labeled as hard. Following a 3:7 sampling ratio between easy and hard queries, we randomly sampled 803, 130, 100, 120, and 120 test instances from these datasets, respectively. The remaining samples were used as support data for ICL. Additionally, we incorporated single-task datasets such as CLEVR~\citep{CLEVR}, FOMI~\citep{FOMI}, and TextOCR~\citep{TextOCR}, adopting the test/support splits specified in VL-ICL Bench~\citep{VL-ICL}. Collectively, these datasets cover a broad spectrum of visual reasoning tasks—including abstract geometry, real-world scenes, charts, graphs, mathematics, spatial relations, counting, attributes and text recognition.

\noindent\textbf{Implementation.} We adopt Gemini-2.0-flash~\citep{Gemini1.5} as the default MLLM policy for ContextNav. The open-source models involved in our experiments, including Phi-3.5V~\citep{Phi}, InternLMX2.5~\citep{internlm}, Qwen2.5-VL~\citep{QwenVL} and embedding models are deployed on an A100 80G GPU, while closed-source models, such as the Gemini series~\citep{GeminiS} and GPT-4o~\citep{gpt4o}, are accessed via APIs on CPU servers. In addition, based on our experimental platform, ContextNav adopts Qwen3-Embedding-4B~\citep{Qwen3Emb} as the textual embedding backbone and the vision encoder of Qwen2.5-VL~\citep{QwenVL} as the visual embedding backbone. By default, the framework employs 8 effective shots for ICL. Unless otherwise specified, ablation experiments are conducted on the MathVision dataset. The prompts involved in the agentic system are documented in Appendix~\ref{appendix_b}, while the definitions of the tool library and the OGG are provided in Appendix~\ref{appendix_c}, and the default embedding model zoos are listed in Appendix~\ref{appendix_d}. In addition, Appendix~\ref{appendix_e} presents case studies of both successful and failure examples. 

\definecolor{darkgreen}{RGB}{0,150,0}
\definecolor{darkred}{RGB}{200,0,0}

\begin{table*}[t]
\small
\caption{\small \textbf{Comparison of downstream accuracies with other baseline methods}, greater values indicate better performance. The bold numbers represent the best accuracy. 'Rand.' is the abbreviation for 'random'. }
\label{tab:quantitative_comp}
\centering
\resizebox{0.96\linewidth}{!}{
\begin{tabular}{lccccccccc}
\toprule
\multirow{2}{*}{\textbf{Methods}} & \multirow{2}{*}{\textbf{BlindTest}} & \multirow{2}{*}{\textbf{RealWorld}} & \multirow{2}{*}{\textbf{CharXiv}}& \multirow{2}{*}{\textbf{GVL}} & \multirow{2}{*}{\textbf{MathVision}} &\multicolumn{3}{c}{\textbf{VL-ICL Bench}}&\multirow{2}{*}{\textbf{Average}}\\ \cmidrule(l){7-9}
&  &  & &  &  & CLEVR & FOMI & TextOCR&\\ 

\midrule
Phi-3.5V-4.2B    &0.402 & 0.292 &0.300   &0.333 & \textbf{0.117}   & 0.425  & 0.070 &  0.715&0.332\\
+Rand. Sample   & 0.339 &  0.254 &  0.250 & 0.308 & 0.092 & 0.395 & 0.070 & 0.685&0.299\\
+VL-ICL   & 0.407 &0.292  & 0.280  & 0.325   &0.100 & 0.435 & \textbf{0.080} & \textbf{0.745}&0.333\\
+MMICES    & 0.399 &   \textbf{0.300}  & 0.300 & 0.317 & 0.083 &0.415 & 0.070 & 0.730 & 0.327 \\
\rowcolor[rgb]{0.80,0.90,0.95} \textbf{+ContextNav}  & \textbf{0.443} & \textbf{0.300} & \textbf{0.310} & \textbf{0.350} &  \textbf{0.117} & \textbf{0.440}  & 0.070 &  0.740 & \textbf{0.346}\\

\midrule
InternLMX2.5-7B    &0.303  &0.262  &0.200   & 0.325 & 0.108  & 0.545 & 0.075  & 0.475&0.287\\
+Rand. Sample   & 0.288 & 0.231 &  0.150 & 0.292 & 0.083 & 0.505 & 0.075 &0.445&0.259\\
+VL-ICL   &0.316    & 0.246  & 0.180 & \textbf{0.342} & 0.092 & \textbf{0.570} & \textbf{0.105} & 0.455 & 0.288\\
+MMICES    & 0.296 &   0.254  & 0.210 &0.333 & 0.092 &0.555 & 0.090 &0.465  & 0.287\\
\rowcolor[rgb]{0.80,0.90,0.95} \textbf{+ContextNav}  &\textbf{0.358} & \textbf{0.277} &\textbf{0.230} & \textbf{0.342} &\textbf{0.142}  & \textbf{0.570} & 0.100 & \textbf{0.480}&\textbf{0.312}\\

\midrule
Qwen2.5-VL-7B    & 0.566 & 0.307 &0.390   &0.342 & 0.217   & 0.820 & 0.045 & 0.835& 0.440\\
+Rand. Sample   & 0.496 & 0.277 & 0.340 & 0.300 &  0.200 & 0.785 & 0.045 &0.820&0.408\\
+VL-ICL   & 0.594 & 0.323 &0.370 & 0.358 &0.233 & 0.900 & \textbf{0.060} &\textbf{0.845}&0.460\\
+MMICES    & 0.606 &   0.315  & 0.390 & 0.342 &0.208 &0.890 & 0.050 & \textbf{0.845}  &0.456\\
\rowcolor[rgb]{0.80,0.90,0.95} \textbf{+ContextNav}  & \textbf{0.645} & \textbf{0.338} &\textbf{0.400}  & \textbf{0.367}   &\textbf{0.250} &  \textbf{0.940} & 0.055 &\textbf{0.845}&\textbf{0.480}\\

\midrule
Gemini-1.5-flash   &0.755  &0.300  &0.490   & 0.475 & 0.483 & 0.535 & 0.120   & 0.880&0.505\\
+Rand. Sample   &  0.733 &  0.285 & 0.440 &0.492 &  0.350 & 0.510 &0.140 & 0.860&0.476\\
+VL-ICL  & 0.775 &0.315 & 0.500 &0.533 & 0.475 & 0.555 & 0.180 &  \textbf{0.890}&0.528 \\
+MMICES    & 0.802 & 0.338   & 0.520 & 0.508 & 0.467 & 0.540 & 0.170 &  0.875& 0.528\\
\rowcolor[rgb]{0.80,0.90,0.95} \textbf{+ContextNav} &\textbf{0.859}  & \textbf{0.369} &  \textbf{0.570} & \textbf{0.575}  & \textbf{0.517} & \textbf{0.575}  & \textbf{0.240} &  \textbf{0.890} &\textbf{0.574}\\

\midrule
Gemini-2.0-flash   &0.761  & 0.308  & 0.510    & 0.508 &0.492 & 0.775  & 0.080 &  0.900& 0.542\\
+Rand. Sample   &0.733 & 0.292 & 0.430 & 0.517 & 0.341 & 0.755 &0.095 & 0.870&0.504\\
+VL-ICL     & 0.773 &   0.331  & 0.490 & 0.550  & 0.483 & 0.810 & 0.145 & 0.905  &0.561\\
+MMICES    & 0.800 &  0.338  & 0.520 &0.525 & 0.467 & 0.795 & 0.140 &  0.885 &0.559\\
\rowcolor[rgb]{0.80,0.90,0.95} \textbf{+ContextNav} &\textbf{0.854} &  \textbf{0.377} & \textbf{0.560} & \textbf{0.600} & \textbf{0.550} &  \textbf{0.825} & \textbf{0.155}  & \textbf{0.910}&\textbf{0.604}\\

\midrule
GPT-4o      & 0.609 & 0.323  & 0.530 & 0.500 & 0.342  & 0.610  & 0.085 & 0.870 &0.484\\
+Rand. Sample   & 0.588  & 0.308 & 0.470 & 0.517  &0.308  &0.635  & 0.095  & 0.860 &0.473\\
+VL-ICL      &  0.631 &  0.338 & 0.540 & 0.558 & 0.333 &  0.650 & 0.140 &  0.890 &0.510\\
+MMICES    &0.649 &    0.353 & 0.550 & 0.542 & 0.316 &0.655 & 0.150 &  0.895 &0.514\\
\rowcolor[rgb]{0.80,0.90,0.95}  \textbf{+ContextNav}  & \textbf{0.672} & \textbf{0.392} & \textbf{0.580} & \textbf{0.608} & \textbf{0.383} & \textbf{0.670}  &\textbf{0.165} & \textbf{0.905} &\textbf{0.547}\\
\bottomrule
\end{tabular}
}
\vspace{-0.12in}
\end{table*}

\begin{figure*}[t]
    \centering
    \includegraphics[width=0.87\linewidth]{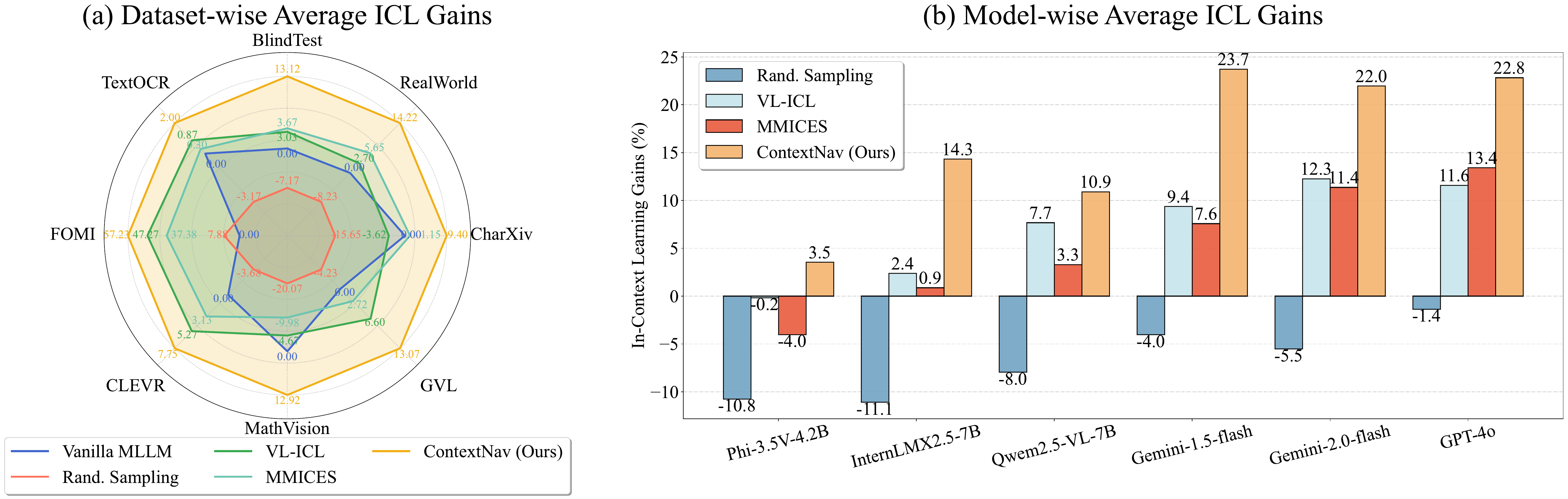}
    \caption{\small Comparison of average ICL gains with baselines. (a) Dataset-wise average gains across 8 datasets. (b) Model-wise average gains across 6 representative MLLMs. 'ICL Gains' represents the percentage improvement of a vanilla MLLM after applying ICL. The complete comparative results are provided in Appendix~\ref{appendix_gains}.}
    \label{fig:comp}
        \vspace{-0.1in}
\end{figure*}

\subsection{Comparison against other methods}
We compare ContextNav with vanilla MLLMs and multimodal ICL baselines under training-free settings, including random sampling, VL-ICL~\citep{VL-ICL}, and MMICES~\citep{CanMLLMs}. The results of VL-ICL and MMICES are obtained from our replications based on their original methodologies. Specifically, the replication of VL-ICL follows the procedure of constructing a manually coarse-filtered candidate pool, followed by random sampling, whereas MMICES adopts a cascaded retrieval process that prioritizes visual retrieval and subsequently applies text-based retrieval. The results in Table~\ref{tab:quantitative_comp} and Figure~\ref{fig:comp} demonstrate that ContextNav rivals or surpasses other carefully designed baselines across nearly all datasets and MLLMs. When the support data involves composite tasks and becomes more complex and noisy (e.g., BlindTest, RealWorld, CharXiv, and MathVision), non-agentic methods are more likely to yield unstable or even degraded performance, whereas ContextNav consistently improves MLLMs’ performance. Overall, ContextNav achieves an average ICL gain of 16.8\% across models and 16.2\% across datasets, substantially outperforming the previous state-of-the-art (7.6\% and 8.2\%, respectively). These results further highlight that ContextNav provides a noise-robust mechanism for exploiting in-context information, delivering consistent and substantial ICL gains across diverse multimodal tasks.

\subsection{Ablation Study}

\begin{table}[t]
\centering
\small
\caption{\textbf{Ablation study on agent policy and core components involved in contextualization.} The bold entries indicate the best results. 'Semantic Noise' and 'Structural Noise' denote the proportions of contextual candidates with the corresponding type of noise, respectively. 'ICL Gains' represents the percentage improvement of a vanilla MLLM after applying ICL. ‘Rand.’ is the abbreviation of random. ‘X-Step TSR’ represents the probability of successfully generating a valid toolchain within 'X' iterations. ‘–’ indicates not applicable. In addition, since Agentic Retrieval enforces semantic alignment, it is disabled when ablating textual and visual retrieval tools to avoid diminishing the significance of the results.}
\label{tab:agent_backbones}
\resizebox{\linewidth}{!}{
\begin{tabular}{lccccc}
\toprule
\textbf{Methods} & \textbf{Semantic Noise $\downarrow$}&\textbf{Structural Noise $\downarrow$} & \textbf{ICL Gains $\uparrow$}  & \textbf{1-Step TSR $\uparrow$}& \textbf{5-Step TSR $\uparrow$} \\
\midrule
\multicolumn{6}{c}{\textit{MLLM policy}}\\
\midrule
Gemini-2.5-pro   & \textbf{0.053}  & 0.084& \textbf{+11.8\%}&  \textbf{1.000} &  \textbf{1.000}  \\
Gemini-2.0-flash &  \textbf{0.053} & 0.084& \textbf{+11.8\%}&   0.995 &  \textbf{1.000} \\
Qwen2.5-VL-7B   &  0.073 &0.107   &+10.1\%&  0.985  &  \textbf{1.000}  \\
Qwen2.5-VL-3B     &  0.080 & 0.139 &+8.4\% &  0.965  & 0.995    \\
\midrule
\multicolumn{6}{c}{\textit{Other Components}}\\
\midrule
w/o Agentic Retrieval (AR)      &  0.171 & 0.090 & +1.6\%      & 0.995& \textbf{1.000}  \\
w/o Structural Alignment   & \textbf{0.053} & 0.573 & +6.7\%   &  0.995 & \textbf{1.000} \\
w/o Textual Retrieval Tools \& AR   & 0.433  & 0.143& -18.7\%   &  0.995& \textbf{1.000}   \\
w/o Visual Retrieval Tools \& AR  & 0.249 & \textbf{0.076}& -3.5\%   &  0.995&\textbf{1.000}  \\
w/o Toolchain Optimization         &0.093 &   0.091 & +5.0\%  & 0.995  & \textbf{1.000}    \\
w/o Operational Grammar Graph    &  -  & -  &   -& 0  &0  \\
\midrule
\rowcolor[rgb]{0.80,0.90,0.95} \textbf{Full (with Gemini-2.0-flash)}&  \textbf{0.053} & 0.084& \textbf{+11.8\%}&   0.995 &  \textbf{1.000} \\
\bottomrule
\end{tabular}
}
\vspace{-0.1in}
\end{table}

\begin{table}[t]
\centering
\caption{\textbf{Ablation study of embedding models.} 'Semantic Noise’ and ‘Structural Noise’ w/o AR \& SA denote the proportions of contextual candidates with corresponding type of noise in the absence of Agentic Retrieval (AR) and Structural Alignment (SA). 'Effective Rate' refers to the proportion of retrieved candidates retained after applying Agentic Retrieval. 'ICL Gains' represents the percentage improvement of a vanilla MLLM after ICL.}
\label{tab:embedding_models}
\resizebox{\linewidth}{!}{
\begin{tabular}{l l c c c  c}
\toprule
\multirow{2}{*}{\textbf{Text Encoder}} & \multirow{2}{*}{\textbf{Vision Encoder}} & \textbf{Semantic Noise $\downarrow$}  & \textbf{Structural Noise $\downarrow$} & \multirow{2}{*}{\textbf{Effective Rate $\uparrow$}} & \multirow{2}{*}{\textbf{ICL Gains $\uparrow$}}   \\
 &  & \textbf{w/o AR \& SA}  &\textbf{w/o AR \& SA}  &  &   \\
\midrule
\multirow{2}{*}{\textbf{Qwen3-Embedding-8B}} 
 & \textbf{Qwen2.5-VL-VisEnc} &  \textbf{0.168} & \textbf{0.581}  &  \textbf{0.765} &  \textbf{+11.8\%} \\ 
  & CLIP-vis   &  0.215 &  0.610 &  0.719 & +10.1\%  \\
 \midrule
\multirow{2}{*}{Qwen3-Embedding-4B} 
 & Qwen2.5-VL-VisEnc &  0.171 & 0.584  &  0.762 &   \textbf{+11.8\%}\\
  & CLIP-vis            & 0.216  &  0.611 &  0.718 &  +10.1\% \\
 \midrule
\multirow{2}{*}{Qwen3-Embedding-0.6B} 
 & Qwen2.5-VL-VisEnc & 0.194  & 0.595  & 0.749  &  \textbf{+11.8\%} \\
  & CLIP-vis         & 0.227
  & 0.623  &  0.721 &  +10.1\% \\
 \midrule
\multirow{2}{*}{CLIP-text} 
 & Qwen2.5-VL-VisEnc &  0.276 &  0.631 &  0.652  &+8.4\% \\
  & CLIP-vis        & 0.311  & 0.659  & 0.606  &+6.7\%\\
\bottomrule
\end{tabular}
}
    \vspace{-0.1in}
\end{table}


\noindent\textbf{Agent policy and core components involved in contextualization.} The upper part of Table \ref{tab:agent_backbones} indicates that the choice of MLLM policy could affect the effectiveness of ContextNav. More capable multimodal policies demonstrate stronger capacity for executing agentic retrieval and structural alignment, thereby reducing both semantic and structural noise and yielding higher ICL gains. In addition, stronger models exhibit improved instruction-following ability, which translates into more reliable toolchain generation and execution. The lower part of the table highlights the contribution of individual modules. The ablation results highlight that textual, visual, and agentic retrieval modules are crucial, as their removal markedly amplifies semantic noise. 
Semantic noise emerges as the dominant factor shaping ICL gains, exerting a stronger impact than other sources of disturbance. At the same time, structural alignment plays a key role in mitigating structural noise, whose influence on ICL, though secondary to semantic noise, remains non-negligible. Moreover, disabling toolchain optimization and restricting the workflow to a single-step determination could yield suboptimal toolchain selection strategies, which consequently limit the effective mitigation of both semantic and structural noise and thereby reduce the ICL gains. In addition, the OGG constitutes the foundation of tool orchestration, and its removal renders the system incapable of executing valid toolchains.

\noindent\textbf{Embedding.} As shown in Table~\ref{tab:embedding_models}, we conduct a manual ablation of the embedding models in ContextNav. For the text encoder, we evaluate the Qwen3-Embedding series at different parameters and the classical CLIP~\citep{clip}. For the vision encoder, we test the language-aligned Qwen2.5-VL visual transformer and CLIP. We observe that embeddings from models with fewer parameters lead to less accurate retrieval, introducing noisy or suboptimal candidates. While noisy candidates are filtered out in Agentic Retrieval, suboptimal ones may remain and still degrade contextual quality, diminishing ICL gains and highlighting the importance of careful embedding selection. We also find that Qwen3-Embedding-8B and 4B yield nearly identical retrieval under the default setting, showing that indiscriminately adopting larger models may bring diminishing returns and unnecessary resource overhead. This finding highlights the practical significance of ContextNav’s resource-aware design, which adaptively selects embedding models according to both the user’s resource usage preferences and the objective hardware constraints.
\begin{figure*}[t]
    \centering
    \includegraphics[width=\linewidth]{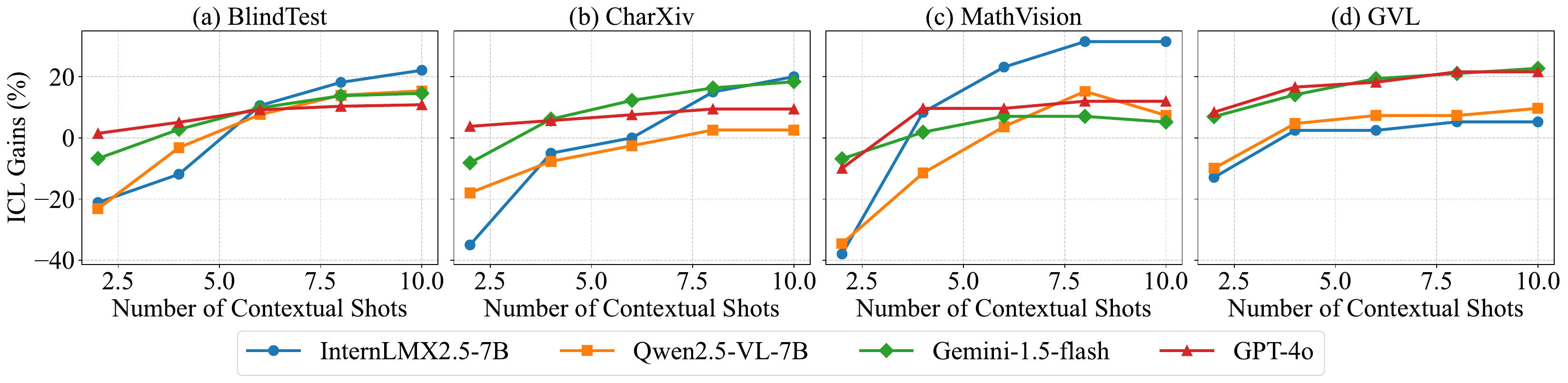}
    \caption{\small \textbf{Ablation study on the number of contextual shots.} 'ICL Gains’ represents the percentage improvement of a vanilla MLLM after ICL.}
    \label{fig:shots}
    \vspace{-0.15in}
\end{figure*}

\noindent\textbf{Number of contextual shots.} As shown in Figure~\ref{fig:shots}, we conduct an ablation study on four MLLMs across four datasets to examine the impact of contextual shot numbers on ICL gains. The results indicate that ICL gains generally improve with more shots but plateaus as the number increases. Smaller models (e.g., Qwen2.5VL-7B and InternLMX2.5-7B) are more sensitive, with too few shots could yield negative gains, whereas larger models (e.g., Gemini-1.5-flash and GPT-4o) exhibit more stable improvements. These results highlight the need for an appropriate choice of shot number: too few degrade performance, while too many add computational cost without commensurate benefit. Accordingly, we set the default to 8 shots in our experiments.



\begin{figure*}[t]
    \centering
    \includegraphics[width=\linewidth]{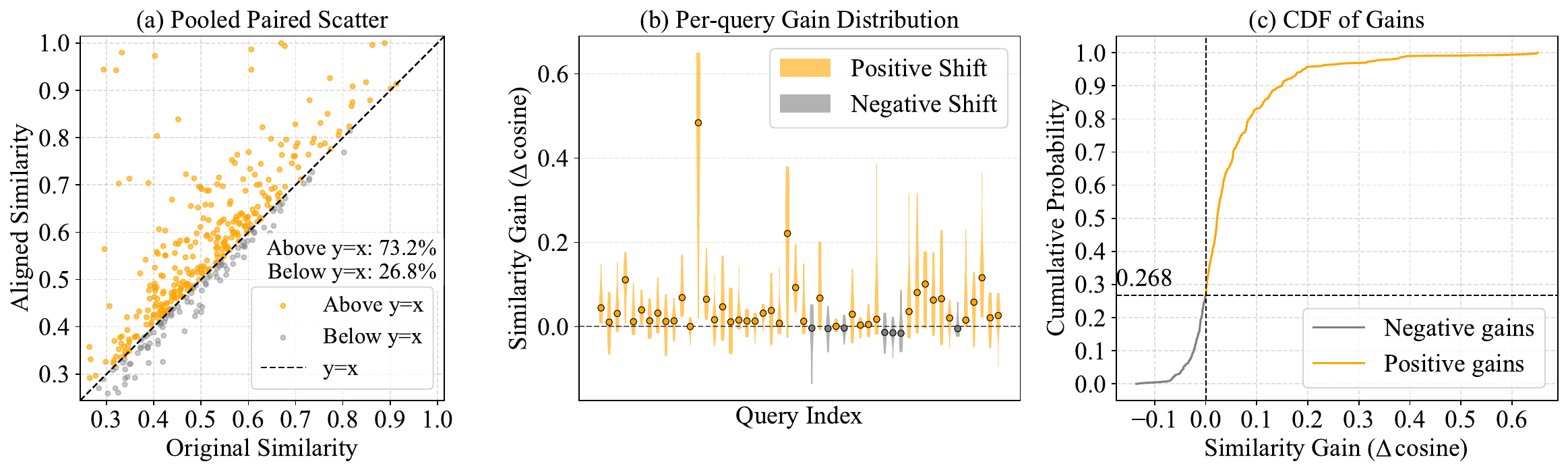}
    \caption{\small \textbf{Effect of Structural Alignment on textual similarity distributions.} (a) Paired scatter between original and aligned similarities. (b) Per-query distributions of similarity gains, with orange denoting positive shifts and gray denoting non-positive shifts, and the points indicate the average gains. (c) Cumulative Distribution Function (CDF) of similarity gains, with reference lines indicating zero‐gain boundaries.}
    \label{fig:sa}
    \vspace{-0.1in}
\end{figure*}

\subsection{Discussion}
\label{sec:dis}
\noindent\textbf{Structural Alignment.} As shown in Figure~\ref{fig:sa}, we analyze the effect of Structural Alignment on textual similarity distributions. We randomly sample 50 queries along with their eight most similar candidates from MATH-Vision dataset and apply structural alignment to them. Panel (a) compares original and refined textual similarities, showing that 73.2\% of the points lie above the diagonal, indicating that aligned generally increases similarity. Panel (b) illustrates per-query distributions of similarity gains, where most queries exhibit a positive shift; although a few negative gains remain, they are marginal in magnitude. Panel (c) presents the cumulative distribution of similarity gains, further confirming that a substantial proportion of candidates benefit from refinement. Overall, these results demonstrate that Structural Alignment mitigates structural discrepancies between queries and candidates, leading to more consistent and semantically aligned contexts.

\noindent\textbf{Negative effects of noisy contexts.} {As shown in Table~\ref{tab:quantitative_comp}, contextualization with random sampling strategies often results in substantial performance degradation. This decline arises because random sampling introduces a large number of query-irrelevant examples, thereby injecting noise into the context. Furthermore, as reported in Table~\ref{tab:agent_backbones}, higher proportions of semantic and structural noise generally correspond to reduced ICL gains. These findings underscore the detrimental impact of noisy contexts in ICL and highlight the necessity of accurate exemplar retrieval. This observation is also consistent with the implicit gradient descent perspective of ICL~\citep{G1,G2,G3}, in which irrelevant or misaligned examples distort the optimization trajectory and hinder generalization.}


\noindent\textbf{Limitations.} Since the agent requires additional inference and tool-execution steps, ContextNav inevitably introduces extra token overhead and system latency, which may limit its applicability in scenarios with stringent real-time requirements. On average, each ICL iteration under the default setting consumes 22.51K tokens and incurs 3.26 seconds of delay. As a complement, Appendix~\ref{appendix_delay} reports the token overhead and system latency when using different MLLMs.

\section{Conclusion}
{In this paper, we introduce ContextNav, the first agentic framework that integrates automated retrieval with human-like curation for multimodal ICL. By combining agentic context management, noise-robust contextualization, and graph-based {workflow} orchestration modules, ContextNav constructs and optimizes noise-resilient contexts within a fully automated workflow, enhancing the multimodal ICL performance of downstream MLLMs. Experiments demonstrate that ContextNav achieves state-of-the-art results across diverse datasets and models, underscoring the potential of agentic workflows for scalable, adaptive, and robust contextualization in multimodal ICL.}




\nocite{ge2024can}
\nocite{ge2025innate}
\nocite{ge2025focusingcontrastiveattentionenhancing}
\nocite{ge2023attack}

\nocite{ge2025famemindframeinterleavedvideoreasoning}
\nocite{ge2025mrfdmultiregionfusiondecoding}

\nocite{fu2025vistawise}
\nocite{fu2025brainvis}
\nocite{fu2024dp}

\nocite{mei2025a1}
\nocite{mei2025surveycontextengineeringlarge}
\nocite{mei2024not}
\nocite{mei2024slang}

\nocite{li2024drs}
\nocite{li2024vulnerability}
\nocite{li2025texture}

\nocite{xiong2025mapping}

\nocite{zhang2024beyond}

\nocite{chen2025haif}

\nocite{ren2025wamo}
\nocite{ren2025sca3d}

\nocite{zhang2024defending}
\nocite{zhang2024tuning}
\nocite{zhang2025improving}
\nocite{zhang2025tokenswap}

\nocite{chai2025causalmace}

\bibliography{iclr2026_conference}
\bibliographystyle{iclr2026_conference}

\appendix

\section{The Use of Large Language Models (LLMs)}
The paper employs LLMs for language polishing.

\section{Full Comparison of ICL Gains}
\label{appendix_gains}
Table~\ref{tab:ICLgains_appendix} presents the complete ICL gain comparison results.
\definecolor{darkgreen}{RGB}{0,150,0}
\definecolor{darkred}{RGB}{200,0,0}

\begin{table*}[h]
\small
\caption{\small \textbf{Comparison of ICL gains (\%) with other baseline methods.} ICL gain represents the percentage improvement of a vanilla MLLM after ICL, greater values indicate better performance.}
\label{tab:ICLgains_appendix}
\centering
\resizebox{\linewidth}{!}{
\begin{tabular}{c|lccccccccc}
\toprule
\textbf{Downstream MLLM} &\textbf{Methods} & \textbf{BlindTest} & \textbf{RealWorld} & \textbf{CharXiv}&\textbf{GVL} & \textbf{MathVision} & CLEVR & FOMI & TextOCR&Mean\\ 
\midrule
 \multirow{4}{*}{Phi-3.5V-4.2B }&Rand. Sample   & -15.6 & -13.0 & -16.7 & -8.1 & -21.4 & -7.1 & 0 & -4.2&-10.8\\
&VL-ICL   & 1.2 & 0 & -6.6 & -2.4 & -14.5 & 2.4 & \textbf{14.3} & 4.2& -0.2\\
&MMICES    & -0.7 & \textbf{2.7} & 0 & -4.8 & -29.1 & -2.4 & 0 & 2.1& -4.0\\
\rowcolor[rgb]{0.80,0.90,0.95} \cellcolor{white}&\textbf{ContextNav} & \textbf{10.2} & \textbf{2.7} & \textbf{3.3} & \textbf{5.1} & 0 & \textbf{3.5} & 0 & \textbf{3.5}&\textbf{3.5}\\
\midrule
 \multirow{4}{*}{InternLMX2.5-7B}  &Rand. Sample   & -5.0 & -11.8 & -25.0 & -10.2 & -23.1 & -7.3 & 0 & -6.3&-11.1\\
&VL-ICL   & 4.3 & -6.1 & -10.0 & \textbf{5.2} & -14.8 & \textbf{4.6} & \textbf{40.0} & -4.2&2.4\\
&MMICES    & -2.3 & -3.1 & 5.0 & 2.5 & -14.8 & 1.8 & 20.0 & -2.1&0.9\\
\rowcolor[rgb]{0.80,0.90,0.95} \cellcolor{white}&\textbf{ContextNav} & \textbf{18.2} & \textbf{5.7} & \textbf{15.0} & \textbf{5.2} & \textbf{31.5} & \textbf{4.6} & 33.3 & \textbf{1.1}&\textbf{14.3}\\
\midrule
\multirow{4}{*}{Qwen2.5-VL-7B}   &Rand. Sample   & -12.4 & -9.8 & -15.0 & -12.3 & -7.8 & -4.5 & 0 & -1.8&-8.0\\
&VL-ICL   & 4.9 & 5.2 & -5.1 & 4.7 & 7.4 & 9.8 & \textbf{33.3} & \textbf{1.2}&7.7\\
&MMICES    & 7.1 & 2.6 & 0 & 0 & -4.1 & 8.5 & 11.1 & \textbf{1.2}&3.3\\
\rowcolor[rgb]{0.80,0.90,0.95} \cellcolor{white}&\textbf{ContextNav} & \textbf{14.0} & \textbf{10.1} & \textbf{2.6} & \textbf{7.3} & \textbf{15.2} & \textbf{14.6} & 22.2 & \textbf{1.2}&\textbf{10.9}\\
\midrule
\multirow{4}{*}{Gemini-1.5-flash}   &Rand. Sample   & -2.9 & -5.0 & -10.2 & 3.6 & -27.5 & -4.7 & 16.7 & -2.3&-4.0\\
&VL-ICL  & 2.6 & 5.0 & 2.0 & 12.2 & -1.7 & 3.7 & 50.0 & \textbf{1.1}&9.4\\
&MMICES    & 6.2 & 12.7 & -3.9 & 6.9 & -3.3 & 0.9 & 41.7 & -0.6&11.4\\
\rowcolor[rgb]{0.80,0.90,0.95} \cellcolor{white}&\textbf{ContextNav} & \textbf{13.8} & \textbf{23.0} & \textbf{16.3} & \textbf{21.1} & \textbf{7.0} & \textbf{7.5} & \textbf{100.0} & \textbf{1.1}&\textbf{23.7}\\
\midrule
\multirow{4}{*}{Gemini-2.0-flash}   &Rand. Sample   & -3.7 & -5.2 & -15.7 & -1.8 & -30.7 & -2.6 & 18.8 & -3.3&-5.5\\
&VL-ICL     & 1.6 & 7.5 & -3.9 & 8.3 & -1.8 & 4.5 & 81.3 & 0.6&16.4\\
&MMICES    & 5.1 & 9.7 & 2.0 & 3.3 & -5.1 & 2.6 & 75.0 & -1.7&11.4\\
\rowcolor[rgb]{0.80,0.90,0.95} \cellcolor{white}&\textbf{ContextNav} & \textbf{12.2} & \textbf{22.4} & \textbf{9.8} & \textbf{18.1} & \textbf{11.8} & \textbf{6.5} & \textbf{93.8} & \textbf{1.1}&\textbf{22.0}\\
\midrule
\multirow{4}{*}{GPT-4o}   &Rand. Sample   & -3.4 & -4.6 & -11.3 & 3.4 & -9.9 & 4.1 & 11.8 & -1.1&-1.4\\
&VL-ICL      & 3.6 & 4.6 & 1.9 & 11.6 & -2.6 & 6.6 & 64.7 & 2.3&11.6\\
&MMICES    & 6.6 & 9.3 & 3.8 & 8.4 & -7.6 & 7.4 & 76.5 & 2.9&13.4\\
\rowcolor[rgb]{0.80,0.90,0.95}   \cellcolor{white}&\textbf{ContextNav} & \textbf{10.3} & \textbf{21.4} & \textbf{9.4} & \textbf{21.6} & \textbf{12.0} & \textbf{9.8} & \textbf{94.1} & \textbf{4.0}&\textbf{22.8}\\
\bottomrule
\end{tabular}
}
\end{table*}

\section{Token Overhead and System Delay}
\label{appendix_delay}
Table~\ref{tab:token_delay} reports the token overhead and system delay of the agent system when using different MLLM policies within ContextNav.
\begin{table}[h]
\centering
\small
\caption{\small Token overhead and system delay of different MLLM policy.}
\label{tab:token_delay}
\begin{tabular}{lcc}
\toprule
\textbf{MLLM Policy} & \textbf{Token Overhead} & \textbf{System Latency (s)} \\
\midrule
Gemini-2.5-pro     & 23.8K & 3.78 \\
Gemini-2.0-flash   & 22.5K & 3.26 \\
Qwen2.5-VL-7B      & 22.6K & 7.38 \\
Qwen2.5-VL-3B      & 22.2K & 5.10 \\
\bottomrule
\end{tabular}
\end{table}

\section{Prompts}
\label{appendix_b}
\begin{tcolorbox}[
  fonttitle = \small\bfseries,
  title=Tool Orchestration Prompt Template,
  colframe=gray!2!black,
  colback=gray!2!white,
  boxrule=1pt,
  boxsep=0pt,
  left=5pt,
  right=5pt,
  fontupper=\setlength{\parskip}{6pt}\footnotesize, 
  halign title = flush center,
]

You are an agent responsible for retrieving information relevant to the user’s query and integrating it into contextual knowledge to assist a multimodal large language model with in-context learning.  

Your available tools are defined as functions with the following descriptions: \{tool\_library\}. From the tool graph \{textualized\_tool\_graph\}, you must select one appropriate toolchain to automate the in-context learning process.

The following are the toolchain(s) you selected in previous steps together with the feedback received for your provided contextual knowledge: \{memory\}. 

\{system\_constraints\}.

Based on your reasoning, decide on the most appropriate toolchain at this step. You \textbf{must} first present your reasoning process, and then output your final decision strictly in the format: 'Toolchain: tool A -> tool B -> $\cdots$ -> tool N.', where the period ``.'' marks the end of the output and \textbf{must not} be omitted.

\end{tcolorbox}

\begin{tcolorbox}[
  fonttitle = \small\bfseries,
  title=System Constraints Template,
  colframe=gray!2!black,
  colback=gray!2!white,
  boxrule=1pt,
  boxsep=0pt,
  left=5pt,
  right=5pt,
  fontupper=\setlength{\parskip}{6pt}\footnotesize, 
  halign title = flush center,
  breakable,
]

There are some criteria you must follow: \{criteria\_and\_retrieval-specification\_Prompt\}.

Please reason these questions and tell me your reasoning results: \{chain\_of\_thought\}. 
\end{tcolorbox}

\begin{tcolorbox}[
  fonttitle = \small\bfseries,
  title=User-defined Criteria and Retrieval-specification Prompt,
  colframe=gray!2!black,
  colback=gray!2!white,
  boxrule=1pt,
  boxsep=0pt,
  left=5pt,
  right=5pt,
  fontupper=\setlength{\parskip}{6pt}\footnotesize, 
  halign title = flush center,
  breakable,
]

1. If you are explicitly instructed that this is your \textbf{first step}, you \textbf{must} select a toolchain that contains 
the tools \texttt{textual\_similarity\_retrieval} and \texttt{visual\_similarity\_retrieval}. However, you cannot stop at these two tools; the complete toolchain must be specified.  

2. If you are not explicitly told that this is your first step, or if you know it is not your first step (e.g., you already selected a toolchain in the previous step), you may select other toolchains at this step.  

3. You must avoid re-selecting any toolchains that have already been chosen in previous steps.  

4. If \textbf{all} toolchains have already been selected in previous steps, then you must disregard the above criteria and instead select a toolchain that includes at least the tools \texttt{textual\_similarity\_retrieval}, \texttt{visual\_similarity\_retrieval}, and \texttt{agentic\_retrieval}.  
\end{tcolorbox}

\begin{tcolorbox}[
  fonttitle = \small\bfseries,
  title=Chain-of-Thought,
  colframe=gray!2!black,
  colback=gray!2!white,
  boxrule=1pt,
  boxsep=0pt,
  left=5pt,
  right=5pt,
  fontupper=\setlength{\parskip}{6pt}\footnotesize, 
  halign title = flush center,
  breakable,
]

1. Is the current step your \textbf{first step}?  

2. If it is not your first step, list the toolchains you have already used in the previous steps.  

3. Reflect on the feedback you received regarding the retrieved context information. Do you think the issues described in the feedback are related to the toolchains you selected in earlier steps?

\end{tcolorbox}

\begin{tcolorbox}[
  fonttitle = \small\bfseries,
  title=General Prompt for Agentic Retrieval,
  colframe=gray!2!black,
  colback=gray!2!white,
  boxrule=1pt,
  boxsep=0pt,
  left=5pt,
  right=5pt,
  fontupper=\setlength{\parskip}{6pt}\footnotesize, 
  halign title = flush center,
  breakable,
]

<image\_query> <image\_ref>

The two images above, together with the following questions, form two image–question pairs.  

Question 1: \{query\_question\}  

Question 2: \{ref\_question\}  

You don’t need to answer the questions. Just decide whether the two pairs share any similarity, either in the images (content) or in the question types (e.g., both ask for counting, scene understanding, etc.).

- If there is any similarity, reply: 'Judgement-YES'.  

- If there is no similarity, reply: 'Judgement-NO' and briefly explain why.  

(Optional: The similarity criterion does not need to be strict, any reasonable overlap counts as similarity.)

\end{tcolorbox}

\begin{tcolorbox}[
  fonttitle = \small\bfseries,
  title=General Prompt for Structural Alignment,
  colframe=gray!2!black,
  colback=gray!2!white,
  boxrule=1pt,
  boxsep=0pt,
  left=5pt,
  right=5pt,
  fontupper=\setlength{\parskip}{6pt}\footnotesize, 
  halign title = flush center,
  breakable,
]

Rewrite the following question in the style of \{query\_question\}. 

Only output the rewritten question, without any explanations or extra text.

Question to rewrite: \{ref\_question\}.

\end{tcolorbox}

\begin{tcolorbox}[
  fonttitle = \small\bfseries,
  title=Embedding Specification Prompt,
  colframe=gray!2!black,
  colback=gray!2!white,
  boxrule=1pt,
  boxsep=0pt,
  left=5pt,
  right=5pt,
  fontupper=\setlength{\parskip}{6pt}\footnotesize, 
  halign title = flush center,
  breakable,
]

From the following options:  

- Text models: \{text\_emb\_model\_zoo\_prompt\}  

- Visual models: \{vis\_emb\_model\_zoo\_prompt\}  

Select one text model and one visual model based on the hardware status \{hardware\_status\}, considering disk and GPU memory usage base on \{resource\_usage\_preference\}.  

\textbf{Output format:}  
\begin{quote}
Text Embedding: text\_model\_id; 

Image Embedding: visual\_model\_id
\end{quote}

\textbf{Restriction:} Do not generate any additional symbols (e.g., \texttt{**}). If a vector database already exists, it is essential to ensure that the chosen embedding model is compatible with it.

\end{tcolorbox}

\begin{tcolorbox}[
  fonttitle = \small\bfseries,
  title=Default General In-Context Learning Template,
  colframe=gray!2!black,
  colback=gray!2!white,
  boxrule=1pt,
  boxsep=0pt,
  left=5pt,
  right=5pt,
  fontupper=\setlength{\parskip}{6pt}\footnotesize, 
  halign title = flush center,
  breakable,
]

<image 1>...<image n> <image\_query> (Option A)

I will provide a series of reference images, each paired with a corresponding question and answer.  
Your task is to \textbf{reflect on these references and summarize the useful information they convey}.  
After all references have been presented, I will then provide one final image with its question.  
Based on your prior reflections, you should give an answer to this final query.  

The $k$-th reference sample is as follows (repeated for $n$ times):  
\begin{itemize}
    \item \textbf{Image $k$:} <image $k$>  (Option B)
    \item \textbf{Question:} \{ref\_question[$k$]\}  
    \item \textbf{Answer:} \{ref\_answer[$k$]\}  
\end{itemize}

Finally, the last query is:  
\begin{itemize}
    \item \textbf{Final Image:} <image\_query> (Option B)
    \item \textbf{Final Question:} \{query\_question\}  
\end{itemize}

Please use your summarized reflections from the reference samples to answer the final question.  

\{feedback\_request\}

\textit{Note: Both option A and option B are acceptable for injecting images. For models with structured interfaces, text and images can be interleaved in a list format. However, when using open-source models, it is empirically common to place all image tokens at the beginning of the prompt, followed by textual instructions. In such cases, if images need to be dynamically inserted within the prompt, a recommended practice is to inject the contexts across multiple dialogue turns.}

\end{tcolorbox}

\begin{tcolorbox}[
  fonttitle = \small\bfseries,
  title=Feedback Request Prompt,
  colframe=gray!2!black,
  colback=gray!2!white,
  boxrule=1pt,
  boxsep=0pt,
  left=5pt,
  right=5pt,
  fontupper=\setlength{\parskip}{6pt}\footnotesize, 
  halign title = flush center,
  breakable,
]

Please evaluate whether the reference samples you received were helpful and sufficiently rich (with the number of shots approximately matching the preset value \{k\}) in solving your final problem. If they were, additionally output "Judgement-Yes"; if not, additionally output "Judgement-No" and, starting with "Feedback:", explain whether the mismatch arose from the text or the image of the reference samples.

\end{tcolorbox}

\section{Tool Graph}
\label{appendix_c}
\subsection{Tool Library}
\begin{table}[H]
\centering
\renewcommand{\arraystretch}{1.2}
\setlength{\tabcolsep}{6pt}
\resizebox{\linewidth}{!}{
\begin{tabular}{p{0.35\linewidth} p{0.65\linewidth}}
\toprule
\textbf{Atom Tools} & \textbf{Description} \\
\midrule
\texttt{get\_query} & Receives the multimodal query (text and image) and initializes the workflow. \\
\midrule
\texttt{get\_hardware\_status} & Monitors computational resources (e.g., GPU memory, disk capacity) to guide resource-aware embedding model selection. \\
\midrule
\texttt{check\_updating} & Detects newly added or modified samples in the context corpus and triggers re-embedding to ensure database synchronization. \\
\midrule
\texttt{matching\_embedding\_models} & Selects appropriate text and vision embedding models from the model zoo, balancing retrieval quality with hardware efficiency. \\
\midrule
\texttt{multimodal\_embedding} & Converts multimodal corpus into vector representations, forming the basis for retrieval in the vector database. \\
\midrule
\texttt{load\_vector\_database} & Builds and/or loads the multimodal vector database. \\
\midrule
\texttt{textual\_similarity\_retrieval} & Retrieves semantically relevant candidates using text embeddings. \\
\midrule
\texttt{visual\_similarity\_retrieval} & Retrieves visually correlated candidates using vision embeddings. \\
\midrule
\texttt{agentic\_retrieval} & Refines the initially retrieved candidates by filtering irrelevant or misleading examples through the agent’s reasoning, mitigating semantic noise. \\
\midrule
\texttt{structural\_alignment} & Reorganizes textual structures of retrieved candidates, reducing structural noise and improving consistency with the query. \\
\bottomrule
\end{tabular}
}
\caption{Descriptions of atom tools in the tool library.}
\label{tab:tool_library}
\end{table}

\subsection{Operational Grammer Graph}

\begin{pybox}
OGG_edges = [
("start", "get_query"),
("get_query", "get_hardware_status"),
("get_query", "check_updating"),
("get_query", "load_vector_database"),
("check_updating", "get_hardware_status"),
("check_updating", "multimodal_embedding"),
("check_updating", "load_vector_database"),
("get_hardware_status", "matching_embedding_models"),
("matching_embedding_models", "multimodal_embedding"),
("multimodal_embedding", "load_vector_database"),
("load_vector_database", "textual_similarity_retrieval"),
("load_vector_database", "visual_similarity_retrieval"),
("textual_similarity_retrieval", "visual_similarity_retrieval"),
("textual_similarity_retrieval", "agentic_retrieval"),
("textual_similarity_retrieval", "structural_alignment"),
("visual_similarity_retrieval", "textual_similarity_retrieval"),
("visual_similarity_retrieval", "agentic_retrieval"),
("visual_similarity_retrieval", "structural_alignment"),
("agentic_retrieval", "structural_alignment"),
("textual_similarity_retrieval", "end"),
("visual_similarity_retrieval", "end"),
("agentic_retrieval", "end"),
("structural_alignment", "end")
]
\end{pybox}

\section{Default Model Zoo}
\label{appendix_d}
\begin{pybox}
text_embedding_model_zoo = [
{"model_id": "Qwen/Qwen3-Embedding-8B",
        "description": "Requires 18 GB of disk space and at least 32 GB of available GPU memory."},
    
{"model_id": "Qwen/Qwen3-Embedding-4B",
        "description": "Requires 9 GB of disk space and at least 16 GB of available GPU memory."},
    
{"model_id": "Qwen/Qwen3-Embedding-0.6B",
        "description": "Requires 2 GB of disk space and at least 8 GB of available GPU memory."},
    
{"model_id": "openai/clip-vit-large-patch14",
        "description": "Requires 2 GB of disk space and at least 8 GB of available GPU memory."}
]

\end{pybox}

\begin{pybox}
vis_embedding_model_zoo = [
{"model_id": "Qwen/Qwen2.5-VL-3B-Instruct",
        "description": "Requires 8 GB of disk space and at least 4 GB of available GPU memory (for ViT)."},  
{"model_id": "openai/clip-vit-large-patch14",
        "description": "Requires 2 GB of disk space and at least 8 GB of available GPU memory."}
]
\end{pybox}

\section{Case Study}
\label{appendix_e}

\subsection{Successful Cases}

\begin{figure*}[h]
    \centering
    \includegraphics[width=\linewidth]{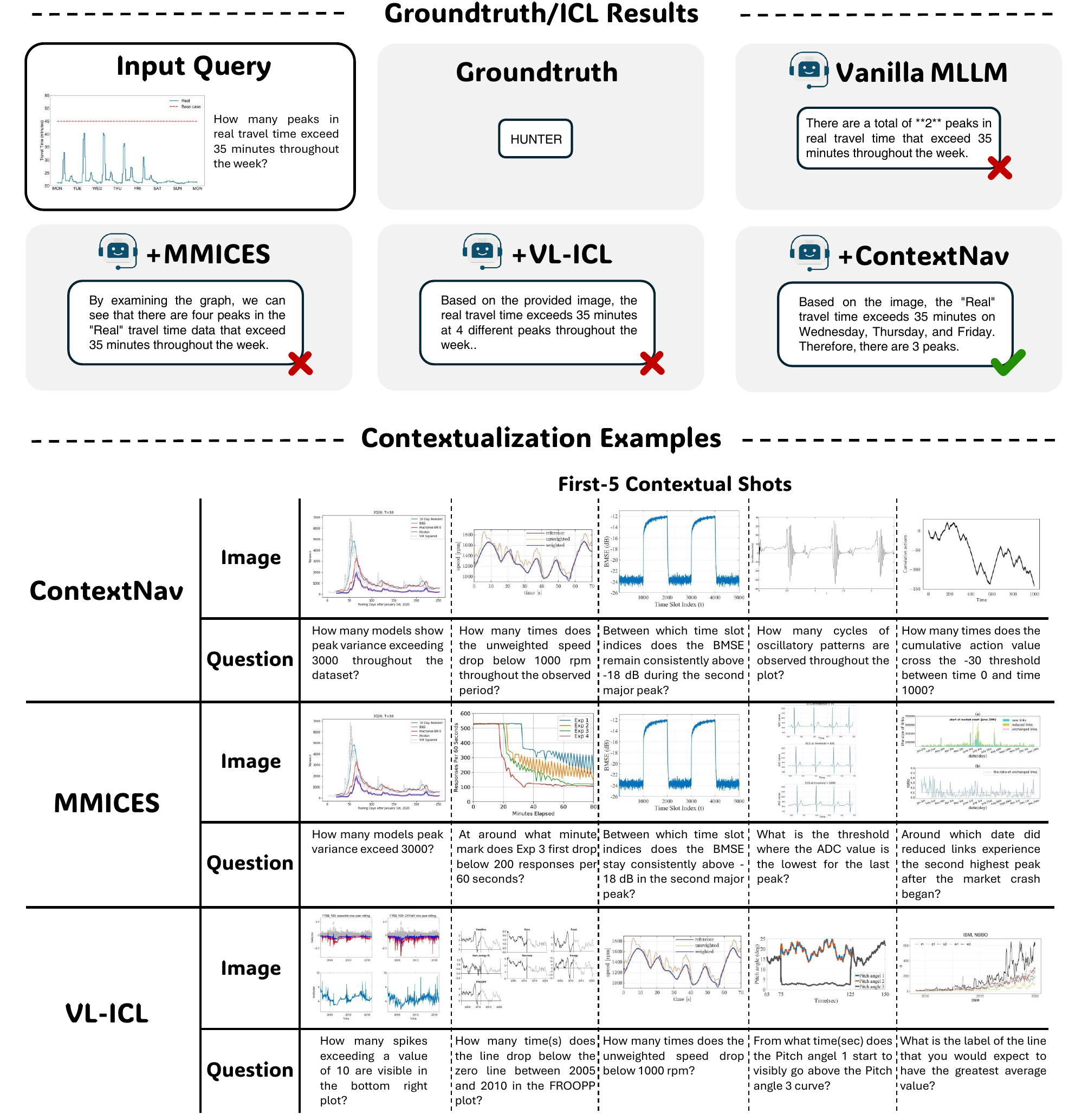}
    \caption{\small Example successful case from CharXiv.}
    \label{fig:success_case_charxiv}
    \vspace{-0.1in}
\end{figure*}

\begin{figure*}[h]
    \centering
    \includegraphics[width=\linewidth]{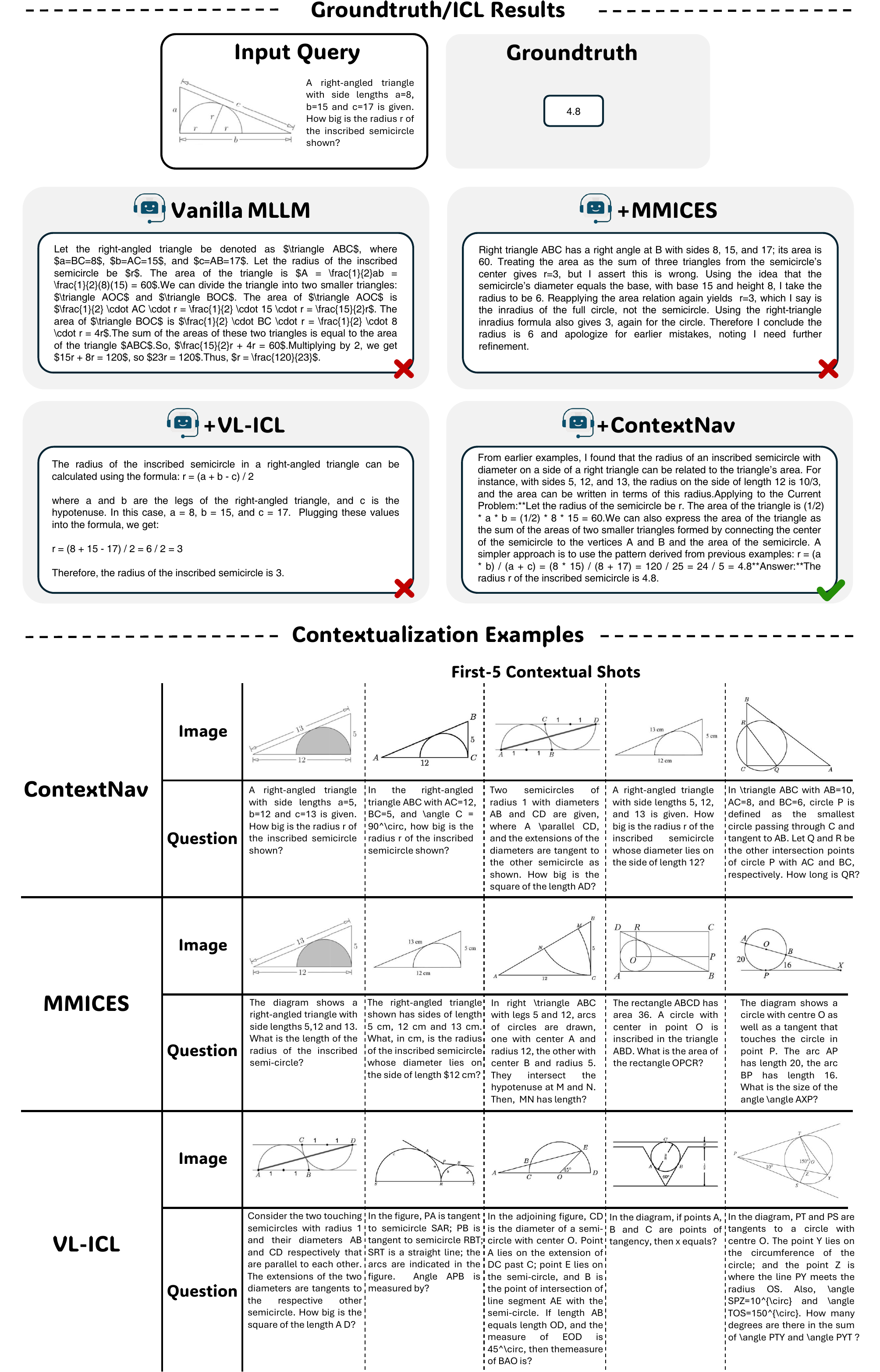}
    \caption{\small Example successful case from MathVision.}
    \label{fig:success_case_mathvision}
    \vspace{-0.1in}
\end{figure*}

We present in Figure~\ref{fig:success_case_charxiv} and Figure~\ref{fig:success_case_mathvision} a comparative case study using Gemini-2.0-flash as the target MLLM, illustrating the differences between ContextNav and alternative approaches in contextualization shot selection and the resulting ICL performance. As the figures demonstrate, ContextNav more effectively identifies semantically relevant contextual shots and aligns their textual structures more closely with the query, thereby enabling the MLLM to conduct ICL more reliably and generate more accurate responses to the input query.

\subsection{Failure Cases}
\begin{figure*}[h]
    \centering
    \includegraphics[width=\linewidth]{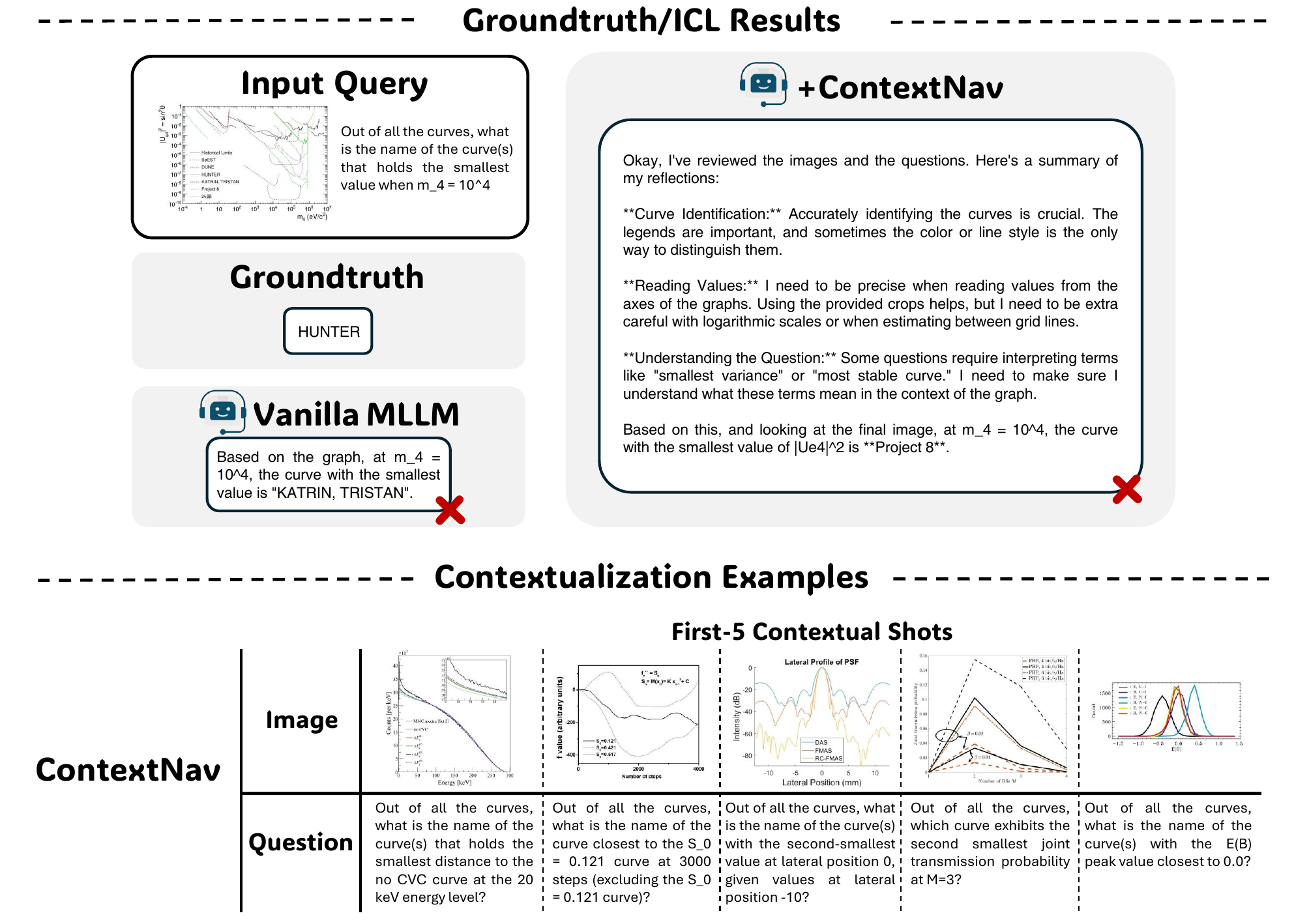}
    \caption{\small Example failure case from CharXiv.}
    \label{fig:fail_case_charxiv}
\end{figure*}

\begin{figure*}[h]
    \centering
    \includegraphics[width=\linewidth]{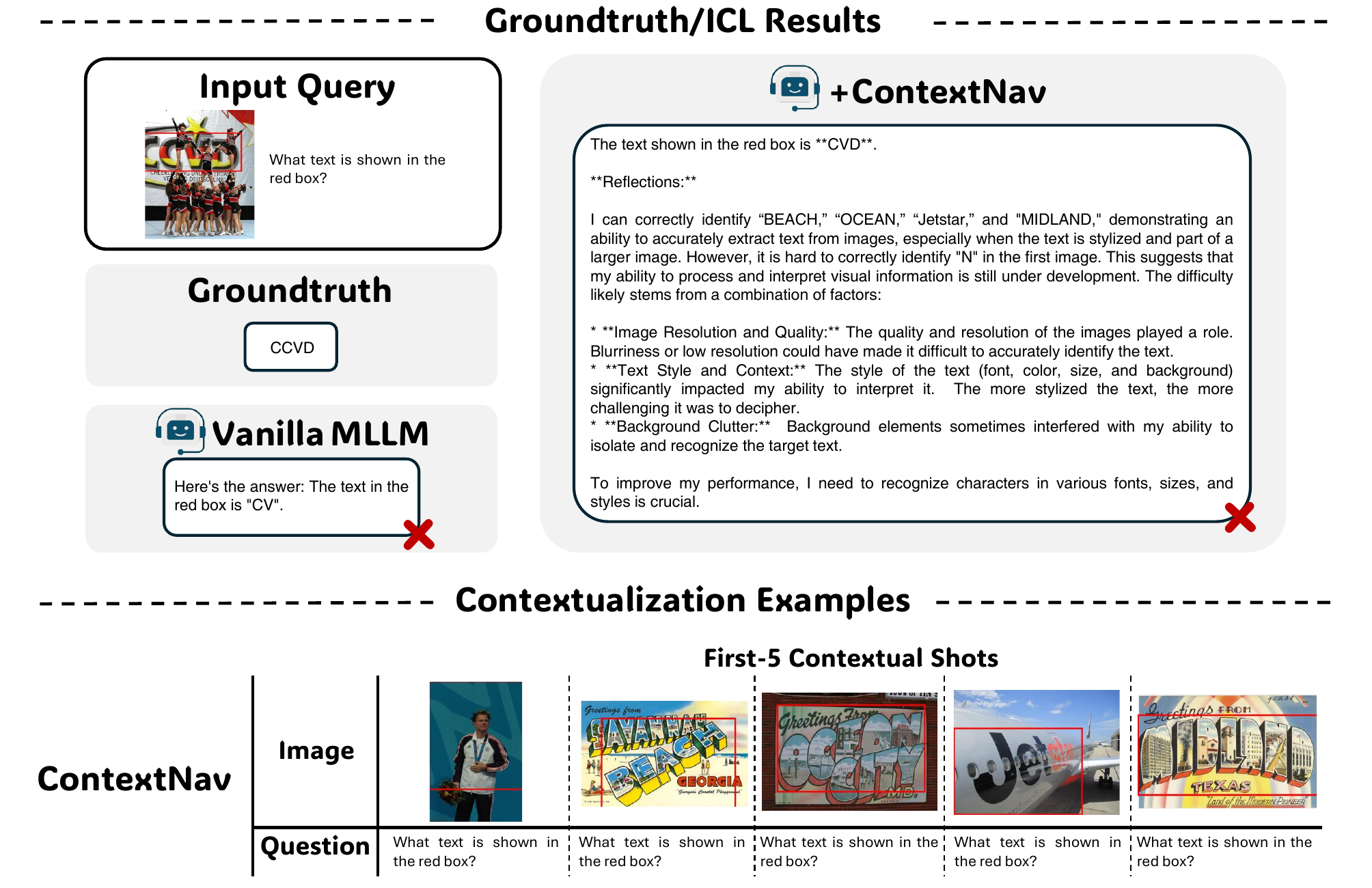}
    \caption{\small Example failure case from TextOCR.}
    \label{fig:fail_case_textocr}
\end{figure*}

We illustrate in Figure~\ref{fig:fail_case_charxiv} and Figure~\ref{fig:fail_case_textocr} representative failure cases of ContextNav with Gemini-2.0-flash as the target MLLM. While ContextNav is generally effective at identifying and supplying the target model with relevant contextual shots, and the model is often able to summarize and reflect upon these contexts, the final predictions are not always correct. We view such cases as reflecting the current limitations of downstream MLLMs’ multimodal reasoning capabilities, which are difficult for ContextNav to overcome.

\end{document}